\documentclass[sigconf]{acmart}

\usepackage{epsfig}
\usepackage{graphicx}
\usepackage{amsmath}

\usepackage{mathtools}
\usepackage{siunitx}
\usepackage[figurename=Figure]{caption}
\usepackage{xspace}
\usepackage{xcolor}
\usepackage{multirow}
\usepackage{wrapfig}
\usepackage{amsthm}
\usepackage{thmtools}
\usepackage{subcaption}
\usepackage{algorithm}      
\usepackage[noend]{algpseudocode} 
\usepackage[title]{appendix}
\usepackage{booktabs}
\usepackage[export]{adjustbox}
\usepackage{mathrsfs}  

\newcommand{\etal}{\textit{et al}.}

\renewcommand{\eqref}[1]{\mbox{Equation~(\ref{#1})}}

\def\Vec#1{{\boldsymbol{#1}}}

\def\Mat#1{{\boldsymbol{#1}}}

\newcolumntype{L}[1]{>{\raggedright\arraybackslash}p{#1}}
\newcolumntype{C}[1]{>{\centering\arraybackslash}p{#1}}
\newcolumntype{R}[1]{>{\raggedleft\arraybackslash}p{#1}}

\acmSubmissionID{245}
\begin{document}

\title{Distance Matters in Human-Object Interaction Detection}


\author{Guangzhi Wang}
\email{guangzhi.wang@u.nus.edu}
\affiliation{
    \institution{
    Institute of Data Science,
    National University of Singapore
    }
    \country{}
}

\author{Yangyang Guo}
\authornote{Corresponding Author.}
\email{guoyang.eric@gmail.com}
\affiliation{
    \institution{
    School of Computing,
    National University of Singapore
    }
    \country{}
}

\author{Yongkang Wong}
\email{yongkang.wong@nus.edu.sg}
\affiliation{
    \institution{
    School of Computing,
    National University of Singapore
    }
    \country{}
}

\author{Mohan Kankanhalli}
\email{mohan@comp.nus.edu.sg}
\affiliation{
    \institution{
    School of Computing,
    National University of Singapore
    }
    \country{}
}


\begin{abstract}
    Human-Object Interaction (HOI) detection has received considerable attention in the context of scene understanding.
    Despite the growing progress on benchmarks,
    we realize that existing methods often perform unsatisfactorily on distant interactions, where the leading causes are two-fold:
    1) Distant interactions are by nature more difficult to recognize than close ones.
    A natural scene often involves multiple humans and objects with intricate spatial relations, making the interaction recognition for distant human-object largely affected by complex visual context.
    2) Insufficient number of distant interactions in benchmark datasets results in under-fitting on these instances.
    To address these problems, in this paper, we propose a novel two-stage method for better handling distant interactions in HOI detection.
    One essential component in our method is a novel Far Near Distance Attention module.
    It enables information propagation between humans and objects, whereby the spatial distance is skillfully taken into consideration.
    Besides, we devise a novel Distance-Aware loss function which leads the model to focus more on distant yet rare interactions. 
    We conduct extensive experiments on two challenging datasets -- HICO-DET and V-COCO.
    The results demonstrate that the proposed method can surpass existing approaches by a large margin, resulting in new state-of-the-art performance.
\end{abstract}

\maketitle
\begin{figure}[t]
    \centering
    \includegraphics[width=1.0\columnwidth]{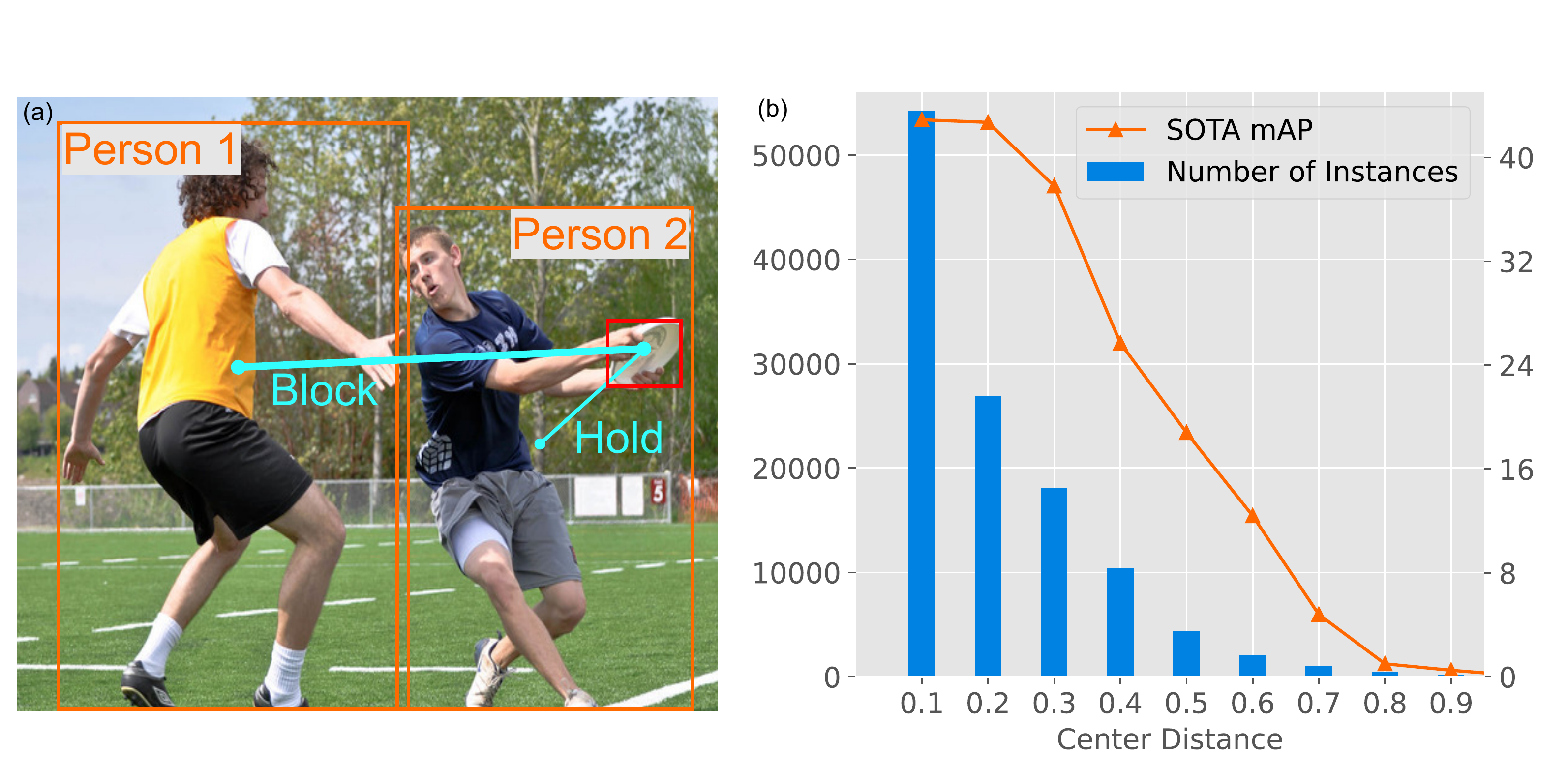}
    \vspace{-1.9em}
    \caption{\label{fig:teaser}
    (a) An example of distance influence for interaction prediction. The distant interaction $\langle$\texttt{person1}, \texttt{block}, \texttt{frisbee}$\rangle$ is harder to predict than the close one $\langle$\texttt{person2}, \texttt{hold}, \texttt{frisbee}$\rangle$.
    (b) Performance variance of state-of-the-art method UPT~\cite{zhang2021upt}
    and instance numbers
    with respect to normalized human-object center distances on the HICO-DET dataset~\cite{HORCNN_Chao2018WACV}.
    It can be seen that distant interactions are relatively sparse and lead to worse results of UPT.
     }
\end{figure}

\section{Introduction}
\vspace{0.5em}
Given a natural image, the task of Human-Object Interaction (HOI) detection is to localize all humans and objects, and recognize the interaction between each human-object pair.
It constitutes an important step towards high-level scene understanding, and has benefited many multimedia applications, including visual question answering~\cite{antol2015vqa, guo2019quantifying}, image captioning~\cite{vinyals2016show_imagecaption} and surveillance event detection~\cite{adam2008robust_eventdetection}.

This task is by its nature challenging, as visual recognition and spatial relation understanding are both required.
Existing efforts can be mainly categorized as two-stage and one-stage methods based on their detection strategy. 
Two-stage methods~\cite{HORCNN_Chao2018WACV, gao2018ican_BMVC18, gao2020drg_ECCV20, xu2019learning_knowledge_CVPR19} initially employ 
an off-the-shelf detector (\textit{e.g.}, Faster R-CNN~\cite{ren2015fasterrcnn}) to detect all humans and objects in the image.
Thereafter, the detected humans and objects are exhaustively paired,
followed by another network for interaction classification. 
By contrast, one-stage methods~\cite{liao2020ppdm_CVPR20, kim2020uniondet_ECCV20, zhong2021glance_GGNet_CVPR21} attempt to solve this problem in a single stage.
In particular, with the success of the Transformer~\cite{transformer} architecture, many approaches~\cite{tamura2021qpic_CVPR21, kim2021hotr_CVPR21, zou2021end_HOITransformer_CVPR21, zhang2021mining_CDN_NIPS21} adopt the attention-based DETR~\cite{carion2020detr} framework in HOI detection and have achieved better performance than their two-stage counterparts.
However, the slow convergence of DETR results in high memory cost and increasing training overhead.
By this reason, Zhang~\textit{et al.}~\cite{zhang2021upt} proposed to revitalize two-stage approaches with the help of DETR.
For this method, the object detection is achieved with a fixed DETR in the first stage, and another attention-like mechanism is adopted for predicting the interactions.

In fact, performing interaction recognition expects accurate understanding about the spatial relation between human and objects, where the distance serves as one fundamental characteristic. 
Nevertheless, predicting interactions with different human-object distances is of distinct difficulties.
Take Fig.~\ref{fig:teaser} (a) as an example, the distant interaction $\langle$\texttt{person1}, \texttt{block}, \texttt{frisbee}$\rangle$ is harder to recognize than $\langle$\texttt{person2}, \texttt{hold}, \texttt{frisbee}$\rangle$, even for us humans.
This observation is further reflected in Fig.~\ref{fig:teaser} (b), where state-of-the-art method UPT~\cite{zhang2021upt} performs much worse on distant interactions.
We attribute this problem to two inherent reasons. 
Firstly, distant interactions are intrinsically more difficult to recognize, owing to the scene complexity.
A natural image often includes multiple humans and objects, while some of them overlap and entangle with each other.
As a result, predicting interaction for distant human-object pairs is largely affected by their noisy context.
Furthermore, distant interactions usually involve small objects, on which the detection backbone struggles due to low-resolution and occlusion.  
Secondly, the number of interactions with respect to human-object center distance demonstrates a long-tail distribution (see Fig.~\ref{fig:teaser} (b)).
Thus, it is unfavorable to fit distant human-object pairs in the training process, resulting in sub-optimal performance.
Nevertheless, to the best of our knowledge, such an important problem remains unexplored by the HOI detection literature.

In this paper, we make contributions from the following two aspects to address the aforementioned problem.
Firstly, we shed light on the two-stage training scheme in HOI detection and propose a novel two-stage method, dubbed Spatially Differentiated Transformer (SDT).
Inspired by~\cite{zhang2021upt}, 
we employ DETR to detect humans and objects (\textit{i.e.} tokens), the representations of which are then enriched via an intra-class diversification module and spatial fusion.
Then,
we design a novel Far-Near Distance Attention (FNDA) mechanism to enable improved modeling for distant token pairs.
Specifically, FNDA allows one token to propagate information with two glances: the first glance involves only far away tokens while the second glance focuses solely on near tokens. 
In this way, the model is able to better attend to distant tokens without the influence of nearby ones and vice versa. 
Thereafter, we pair the obtained features of each human and object, and employ another self-attention module for iterative context aggregation.
At last, each human-object pair is classified into candidate interactions.

Furthermore, as the number of interactions manifests a long-tail distribution with respect to human-object distances, we therefore design a novel Distance-Aware (DA) loss function to re-weight each human-object pair during training.
In particular, DA loss adaptively adjusts the weights for each human-object pair, wherein relatively higher and lower weights are assigned to distant and close interactions, respectively.
It is expected that with our DA loss, distant interactions are treated with more importance, thereby alleviating the effect of dominance of close interactions during training.

To validate the effectiveness of our proposed method, we conduct extensive experiments on two challenging benchmarks, namely HICO-DET~\cite{HORCNN_Chao2018WACV} and V-COCO~\cite{gupta2015visual_VCOCO}. 
The results show that our method outperforms existing approaches by a large margin, resulting in new state-of-the-art performance on the two benchmarks. 
Besides, additional analysis and visualizations further demonstrate that our method is able to better model distant interactions.

To summarize, the contributions of this paper are three-fold:
\begin{itemize}
    \item We propose a novel two-stage method -- Spatially Differentiated Transformer for HOI detection, wherein a Far-Near Distance Attention (FNDA) is devised to effectively model distant interactions. 
    \item To balance the learning of interactions with different human-object distances,
    we design a novel Distance-Aware (DA) loss function to dynamically adjust the weight of 
    each human-object pair according to their center distance.
    \item Extensive experiments on two benchmarks demonstrate that the proposed method surpasses existing approaches by a large margin, achieving new state-of-the-art results\footnote{Code available: \url{https://github.com/daoyuan98/SDT-HOI}.}.
\end{itemize}
\section{Related Work}
\vspace{0.5em}
\subsection{Human-Object Interaction Detection}
\vspace{0.5em}
Existing approaches for HOI detection can be mainly categorized as two-stage and one-stage methods, according to their detection strategy.
Two-stage methods firstly adopt an off-the-shelf detector (like Faster-RCNN~\cite{ren2015fasterrcnn}) to detect all humans and objects in the image.
Afterwards, the humans and objects are exhaustively paired and fed into a downstream network for interaction recognition. 
These approaches mostly focus on improving the interaction recognition capacity of the downstream network.
Early work~\cite{HORCNN_Chao2018WACV, shen2018scaling_WACV} often leverages the visual appearance and spatial relation to identify the interactions. 
In addition, more features such as human pose~\cite{gupta2019nofrills_ICCV19, wan2019poseaware}, gaze~\cite{xu2019interact_iHOI_TMM}, 3D representations~\cite{li2020djrn_CVPR20} and word embeddings~\cite{kim2021acp++_TIP, liu2020consnet_CVPR20},
have also been exploited.
Different from these methods, some studies utilize the graph structure for information propagation between the detected humans and objects.
For example, \cite{qi2018learning_GPNN_ECCV18} builds a fully-connected graph while \cite{gao2020drg_ECCV20, zhang2021spatially_SCG_ICCV21} apply a bipartite graph for humans and objects.
Despite their effectiveness, these methods usually suffer from sub-optimal detection performance~\cite{gao2020drg_ECCV20, li2020hoi_IDN, zhang2021spatially_SCG_ICCV21}. 

By contrast, one-stage methods tackle the task in an end-to-end manner.
Early one-stage approaches often detect the interaction points~\cite{liao2020ppdm_CVPR20, wang2020learning_IPNet_CVPR20} or the human-object union regions~\cite{kim2020uniondet_ECCV20} as interaction clues.
Recently, with the success of DETR in the object detection domain, some studies attempt to employ this Transformer-based architecture for HOI detection.
For instance, \cite{tamura2021qpic_CVPR21, zou2021end_HOITransformer_CVPR21, kim2021hotr_CVPR21} append more heads on the decoder to recognize the interaction.
\cite{chen2021reformulating_ASNet_CVPR21} applies another decoder for decoupled object detection and interaction recognition, while \cite{zhang2021mining_CDN_NIPS21} performs the two steps in a cascaded way.

\begin{figure*}
    \centering
    \vspace{1em}
    \includegraphics[width=1.0\textwidth]{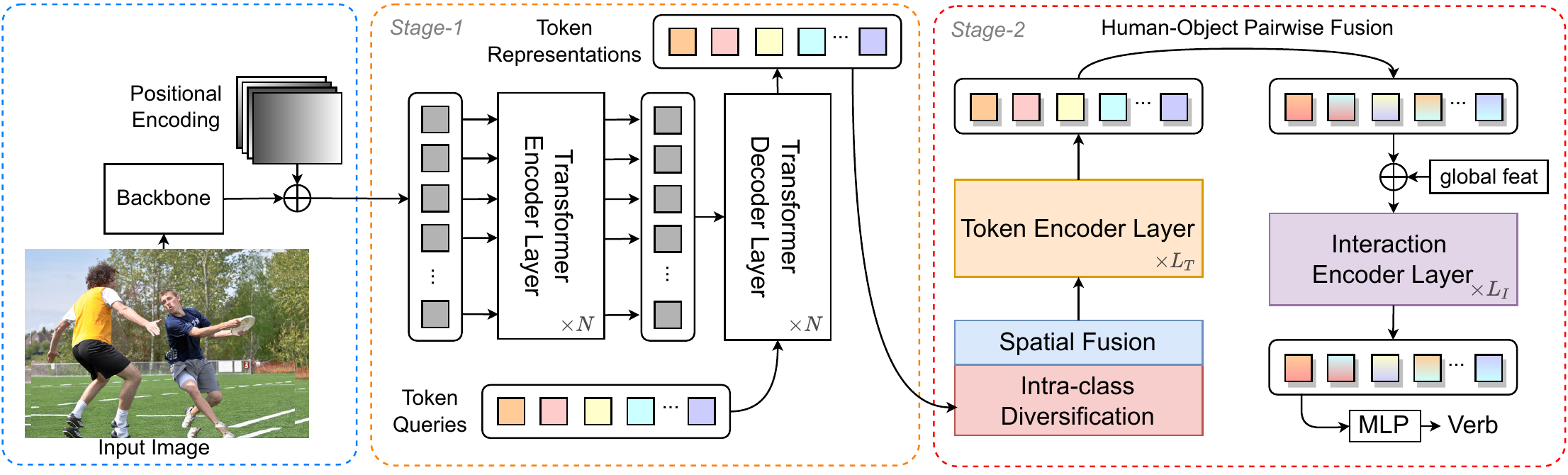}
    \vspace{-1.2em}
    \caption{
    Illustration of the proposed Spatially Differentiated Transformer (SDT).
    Given a natural image, a backbone network is employed to extract a set of visual features. After added with the positional encodings, we adopt DETR~\cite{carion2020detr} to detect all tokens, \textit{i.e.}, humans and objects.
    The token representations are fed into an Intra-Class Diversification module and a spatial fusion module to enrich the interaction information. 
    Then, we employ $L_T$ Token Encoder Layers to propagate information between tokens.
    After that, each human-object pair is fused with the global visual feature, and inputted into $L_I$ Interaction Encoder Layers for iterative context aggregation.
    Finally, an Multi-Layer Perceptron (MLP) is used for interaction prediction. 
    }\label{fig:framework}
\end{figure*}

\subsection{Single- and Cross-Modal Attention}
\vspace{0.5em}
The past few years have witnessed the rapid development of attention mechanism due to its effectiveness in a variety of tasks.  
Among the dedicated efforts, the Multi-Head Self-Attention (MHSA)~\cite{transformer} has recently attracted increasing research interest.
In addition to its wide application in the language domain, some researchers attempt to apply MHSA to the vision tasks, such as image classification~\cite{VIT}, semantic segmentation~\cite{SwinTransformer}, point cloud analysis~\cite{fan21p4transformer,fan2022point} and video understanding~\cite{AttentionVideo}.
Besides, some studies have been conducted to improve the efficiency of the attention mechanism.
For example, \cite{wang2021crossformer, chen2022regionvit, guo2019attentive} present to reduce the computational cost or model bi-level relations by splitting attention into two groups of different receptive fields.
Qin~\etal~\cite{qin2022cosformer} decreases the quadratic computation complexity by eliminating the softmax layer.
Other methods also consider the relative position between inputs by taking it as learned relative bias embedding~\cite{shaw2018RPB} or part of the attention weights~\cite{SwinTransformer, wang2021crossformer}.
While above methods focus only on single modality information, the effectiveness of the attention mechanism on cross-modality tasks has also been extensively studied.
One typical operation is to replace the query matrix with the features from another modality,  and the cross-modality information exchange is accordingly enabled.
It demonstrates improved results on various multi-modal tasks, including  visual question answering~\cite{liu2020bridging}, image-text matching~\cite{liu2019focus, wu2019learning},
zero-shot learning~\cite{xu2021relation} 
and audio-visual active speaker detection~\cite{tao2021someone}.

\section{Methodology}
\vspace{0.5em}

\subsection{Preliminary}
\vspace{0.5em}
\noindent\textbf{Method Intuition.}
Human-Object Interaction (HOI) detection is to detect and predict a set of  $\langle\texttt{human}, \texttt{verb}, \texttt{object}\rangle$ interaction triplets in an image.
In this work, we attempt to address the problem that distant human-object often lead to inferior interaction recognition results.
In particular, we propose a novel two-stage method named Spatially Differentiated Transformer (SDT), which is designed to flexibly model both distant and close interactions. 
As shown in Fig.~\ref{fig:framework}, our method is composed of two stages: The first stage detects human and object (token) with an object detector, followed by the second stage of interaction recognition on the detected tokens.

\vspace{0.5em}
\noindent\textbf{Stage 1: Token Detection.}
Inspired by~\cite{zhang2021upt}, we perform token detection with a Transformer-based DETR~\cite{carion2020detr},
which demonstrates superior results in object detection.
In this step, an input image is first fed into a backbone network, \textit{e.g.}, ResNet-50~\cite{he2016deep}, to obtain the visual representation $\Vec{g} \in \mathbb{R}^{d}$.
With the addition of positional encodings, the visual features are then inputted to $N$ Transformer encoder layers and $N$ Transformer decoder layers sequentially.
Thereafter, we have a set of token representations, corresponding to the token queries inputted to the decoder layers.
Finally, these token representations are fed into two separate Multi-Layer Perceptrons (MLPs) for object classification and bounding box regression, respectively.

\vspace{0.5em}
\noindent\textbf{Stage 2: Interaction Recognition.}
In the second stage, we filter the tokens according to their confidence scores, and keep $n$ most confident ones.
Each token is represented as a feature vector $\Vec{t}_i \in \mathbb{R}^{d}$, normalized bounding box $\Vec{b}_i \in \mathbb{R}^{4}$, confidence score $s_i$, and class $c_i$, where $i = 1,...,n$. 
With the detected tokens, we first post-process these tokens to make them more compatible with HOI detection. 
Then, these tokens are enabled to propagate information with distance discrimination by our Far-Near Distance Attention.
Afterwards, we obtain the interaction representation by combing each human and object tokens with the global context.
Finally, the interaction representations are fed into an MLP for final prediction.
In the rest of this section, we will sequentially elaborate the token post-processing step, the Far-Near Distance Attention, our Distance-Aware loss function, and training and inference procedures.

\subsection{Token Post-Processing}
\vspace{0.5em}
The token representations $\{\Vec{t}_i\}_{i=1}^n$ obtained from DETR contains discriminative information for classification.
However, they can be of low quality and oblivious to the spatial relations with other tokens.
Therefore, we first post-process these tokens to enrich their representation for better interaction recognition.

\vspace{0.5em}
\noindent\textbf{Intra-Class Diversification.}
It is reasonable that the token representations are often of limited diversity.
In particular, tokens representing small size objects often suffer from occlusion and low-resolution due to the scene complexity, and is thus less informative.
To this end, we propose an Intra-Class Diversification (ICD) module to enrich the token representations with the help of other instances from the same class.   

The key to our ICD module is an object-wise memory, which stores token representations of high confidence score for each class.
During training, given a token with representation $\Vec{t}_i$, we take it as a query and randomly sample $l$ features $\{\Vec{t}_k\}_{k=1}^{l}$ from the memory cell corresponding to its class for key and value computation. 
Then, the intra-class diversification is implemented through the cross-attention mechanism~\cite{transformer}:
\begin{equation}
    \left\{
    \begin{array}{lcl}
    \Mat{A}_{ik} &=& \frac{\Vec{t}_i\Mat{W}_Q\Mat{W}_K^{T}\Vec{t}^{T}_k}{\sqrt{d}} \vspace{0.6em} \\
    \Mat{B}_{ik} &=& \frac{e^{\Mat{A}_{ik}}}{\sum_{k'=1}^{l}e^{\Mat{A}_{ik'}}} \vspace{0.6em} \\
    \tilde{\Vec{t}}_i &\leftarrow& \Vec{t}_i + \sum_{k'=1}^{l}\Mat{B}_{ik'}\Vec{t}_{k'}\Mat{W}_V \vspace{0.6em}
    \end{array}
    \right.
\end{equation}
where $\Mat{W}_Q, \Mat{W}_K, \Mat{W}_V \in \mathbb{R}^{d \times d}$ are learnable transformation matrices.
This mechanism allows each token to adaptively aggregate representation from high quality tokens in the same class, thereby improving token diversity and enhancing model's generalizability.

\vspace{0.5em}
\noindent\textbf{Spatial Fusion.}
Besides the visual representation, the bounding box of a detected token also serves as a key attribute for describing the spatial information. 
For understanding the relation between two tokens, it is more important to focus their relative spatial relations.
Thus, inspired by~\cite{zhang2021spatially_SCG_ICCV21, zhang2021upt}, we first obtain the pairwise spatial relation $\Vec{p}_i = f(\Vec{b}_i, \{\Vec{b}_j\}_{j=1}^{n})$, which represents the spatial relation between the $i$-th and all the other tokens\footnote{We detail the computation of 
$f$ in supplementary material due to space limit.}.
The spatial relation vector $\Vec{p}_i$ can be regarded as a high-level positional embedding, which better represents the detected tokens.
Afterwards, we fuse the pairwise spatial relation $\Vec{p}_i$ into the token representations:
\begin{equation}
    \hat{\Vec{t}}_i = \mathsf{FFN}([\tilde{\Vec{t}}_i;\Vec{p}_i]),
\end{equation}
where $\mathsf{FFN}$ denotes a feed forward network.
With the combination of spatial relations, the processed tokens are more informative, and thus more compatible for HOI detection.

\subsection{Far-Near Distance Attention}
\vspace{0.5em}

\begin{figure}
    \centering
    \includegraphics[width=0.88\columnwidth]{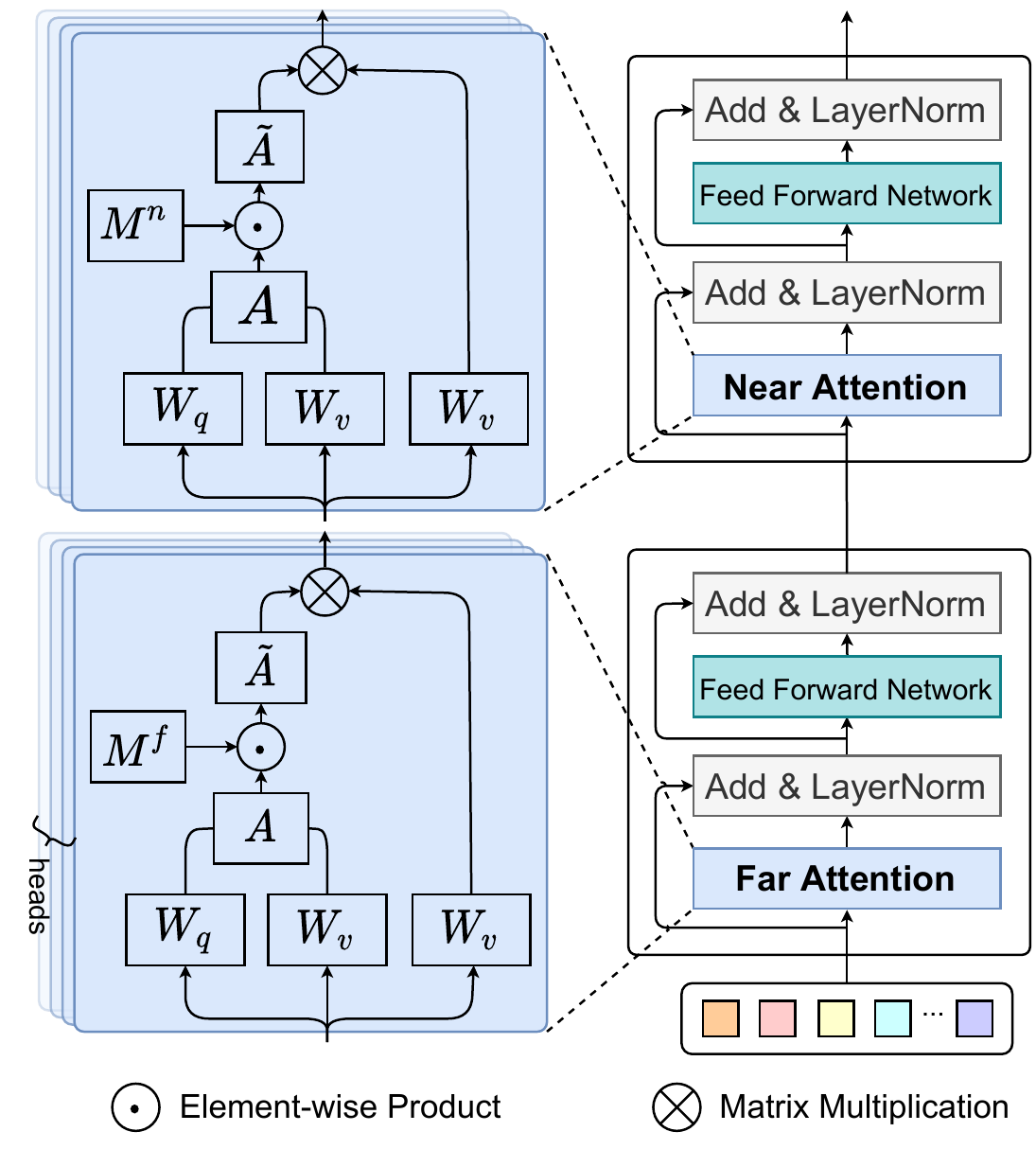}
    \vspace{-1em}
    \caption{Each token encoder layer contains two blocks, 
    where
    far attention mask $\Mat{M}^f$ and near attention mask $\Mat{M}^n$ are alternatively applied.
    Note that the softmax operation is folded for clarity.
    }\label{fig:fnda}
\end{figure}

After fusing the spatial embeddings, one common way to propagate information between tokens is to leverage the Multi-Head Self-Attention (MHSA) mechanism in Transformer~\cite{transformer}.
Nonetheless, it is sub-optimal to directly apply MHSA for HOI detection.
Broadly speaking, one human can interact with multiple objects and vice versa, making MHSA struggle on dense scenes.
Furthermore, interaction prediction for tokens with far and near distances is often of distinct difficulties.
Current MHSA tackles all token-pairs without discrimination, which may unexpectedly impede the modeling for distant relations.

To this end, we design a novel Far-Near Distance Attention (FNDA) mechanism, which models the relation between tokens in two alternative steps, according to their center distance.
Specifically, given the bounding boxes of the detected tokens, we first compute $L_2$ center distance for each token pair, and obtain the pairwise distance matrix $\Mat{D} \in \mathbb{R}^{n \times n}$.
Afterwards, we compute two attention masks $\Mat{M}^f$ and $\Mat{M}^n$ as follows:
\begin{equation}
\begin{split}
    \Mat{M}^f_{ij} &= \left \{
    \begin{array}{ll}
    1,  &   \Mat{D}_{ij} > Med(\Vec{d}_i)\ or\ i=j \\
    0,  &   otherwise  \\
    \end{array}
    \right. \\
    \Mat{M}^n_{ij} &= \left \{
    \begin{array}{ll}
    1,  &   i=j \\
    1 - \Mat{M}^f_{ij}, & otherwise
    \end{array},
    \right.
\end{split}
\end{equation}
where $Med(\Vec{d}_i)$ denotes the median of the $i$-th row in matrix $\Mat{D}$, which serves as the threshold for differentiating far and near distances.
In this way, $\Mat{M}^f$ allows the model to attend to far tokens only, which prohibits the intervention from near ones, and vice versa\footnote{Note that the diagonal elements are not masked in $\Mat{M}^f$ and $\Mat{M}^n$ so that one token can always attend to itself.}. 
Then, we perform the MHSA operation with the distance-guided masks:
\begin{equation}
    \left\{
    \begin{array}{lcl}
    \Mat{A}_{ij}  &=& \frac{\hat{\Vec{t}}_i\Mat{W'}_Q\Mat{W'}_K^{T}\Vec{\hat{t}}_j^T}{\sqrt{d}} \vspace{0.6em}\\
    
    \tilde{\Mat{A}} &=& \Mat{A} \odot \Mat{M}^{f|n} \vspace{0.6em} \\
    
    {\Mat{B}_{ik}} &=& \frac{e^{\tilde{\Mat{A}}_{ik}}}{\sum_{k'=1}^{n}e^{\Mat{\tilde{A}}_{ik'}}}  \vspace{0.6em}\\
    
    \tilde{\Vec{t}_i} &\leftarrow& \tilde{\Vec{t}_i} + \sum_{k'=1}^{k}\Mat{B}_{ik'}\Vec{t}_{k'}\Mat{W'}_V \vspace{0.6em}
    \end{array}
    \right.
\end{equation}
where $\odot$ denotes element-wise product, $\Mat{W'}_Q$, $\Mat{W'}_K$, $\Mat{W'}_V \in \mathbb{R}^{d \times d}$ are learnable transformation matrices , $\Mat{M}^{f|n}$ indicates either $\Mat{M}^{f}$ or $\Mat{M}^{n}$.
In our model, $\Mat{M}^f$ and $\Mat{M}^n$ are alternatively applied, so that each token can iteratively update their representations with tokens of different distances,
thereby improving the model's capacity for distant relation modeling.
It is worth noting that our FNDA can replace MHSA in the Transformer encoder layer with few efforts, resulting in our token encoder layer.
The detailed operation is illustrated in Fig.~\ref{fig:fnda}.

\vspace{0.5em}
\noindent\textbf{Iterative Context Aggregation.}
Thereafter, we pair each human token and object token, and take them as the initial interaction representation for each candidate pair. 
We also fuse the global context feature $\Vec{g}$ into the representations, which can provide contextual clues for interaction recognition:
\begin{equation}
    \Vec{h}_{ij} = [\Vec{\tilde{t}}_i;\Vec{\tilde{t}_j}] \oplus \mathsf{FFN}(\Vec{g}),
\end{equation}
where $\oplus$ denotes element-wise addition, $i$ implies the index for human and $i \neq j$.
We then employ $L_I$ interaction encoder layers to perform self-attention~\cite{transformer} to iteratively update the interaction representations with the global context.
After that, an MLP is used to predict the final interactions for each human-object pair.

\subsection{Distance-Aware Loss}
\vspace{0.5em}
Following previous work \cite{HORCNN_Chao2018WACV, zhang2021spatially_SCG_ICCV21, zhang2021mining_CDN_NIPS21, liu2020consnet_CVPR20}, we formulate interaction recognition with an multi-label classification objective, since there can be multiple interactions (\textit{e.g.}, \texttt{read} \texttt{book} and \texttt{hold} \texttt{book}) within one human-object pair.
In this way, the full model can be optimized with the Binary Cross Entropy (BCE) loss.
However, distant interactions often manifest rare in benchmark datasets, which leads to underfitting on these interactions.
Therefore, we propose a Distance-Aware (DA) loss to adaptively assign higher weights and lower weights to distant and close interactions, respectively.
In formulation, the interaction recognition model is optimized with:
\vspace{0.5em}
\begin{equation}
    \mathcal{L}_{DA} = \sum_{i,j\ i \neq j} w_{ij} \sum_{c\in C} \
    \Vec{y}_{ij}^clog\Vec{\delta}_{ij}^c +
    (1 - \Vec{y}_{ij}^c)log(1 - \Vec{\delta}_{ij}^c), 
\end{equation}

\noindent where $\Vec{\delta}_{ij}= \sigma(\mathsf{MLP}(\Vec{h}_{ij}))$ represents the verb scores transformed by the sigmoid function $\sigma(\cdot)$, $C$ is the number of classes and $\Vec{y}_{ij}^c$ indicates the ground-truth label of class $c$.
We implement $w_{ij}$ as, 
\begin{equation}
    w_{ij} = \sigma(\alpha \cdot \Mat{D}_{ij} + \beta),
\end{equation}
where $\Mat{D}_{ij}$ is the $L_2$ center distance between the $i$-th human and the $j$-th object, and $\alpha$ and $\beta$ are both learnable parameters. 
In this way, relatively higher weights are assigned to more distant interactions, so as to increase their importance during training.
Notably, the proposed DA loss acts on each instance.

\subsection{Training and Inference}
\vspace{0.5em}
\noindent\textbf{Training.}
We first train the object detector, and then freeze it to train the interaction recognition model.
Pertaining to the latter training phase, focal loss~\cite{lin2017focal} has been proven effective to tackle the class-imbalance problem, which is shown to be influential to the performance in HOI detection~\cite{hou2021affordance_ATL_CVPR21, zhang2021spatially_SCG_ICCV21, zhang2021upt}.
Therefore, we combine our DA loss with the focal loss to optimize our model.
Besides, since our method breaks HOI detection into two stages, after detection, infeasible verb-object combinations can be filtered out in advance.

\vspace{0.2em}
\noindent\textbf{Inference.}
During inference,
the final interaction score $\Vec{z}$ is calculated as the multiplication of human confidence score $s_i$, object confidence score $s_j$ and their verb score $\Vec{\delta}_{ij}^c$:
\vspace{0.2em}
\begin{equation}\label{eq:interaction_score}
    \Vec{z}_{ij}^c = (s_i)^{\lambda} \cdot (s_j)^{\lambda} \cdot \Vec{\delta}_{ij}^c,
\end{equation}

\noindent where $s_i$ and $s_j$ are obtained from DETR and $\lambda \geq 1$ is a constant to suppress overconfident objects~\cite{zhang2021spatially_SCG_ICCV21, zhang2021upt}.
Infeasible verb-object combinations are also removed according to the object label output by DETR.
\section{Experiments}
\vspace{0.5em}
\subsection{Experimental Setup}
\vspace{0.5em}
\noindent\textbf{Datasets.} 
We conducted experiments on two benchmark datasets, namely HICO-DET~\cite{HORCNN_Chao2018WACV} and V-COCO~\cite{gupta2015visual_VCOCO}.
\textbf{HICO-DET} involves 80 COCO objects and 117 verb classes, resulting in a total of 600 interaction classes.
There are 38,118 and 9,658 images for training and test in this dataset, respectively.
\textbf{V-COCO} is built upon the MS-COCO~\cite{lin2014MSCOCO} dataset. 
It covers 24 action categories with 80 COCO objects, and contains 2,533, 2,867 and 4,946 images for training, validation and test, respectively.

\vspace{0.5em}
\noindent\textbf{Evaluation Protocol.}
We adopted mean Average Precision (mAP) as the evaluation metric. 
A detection result is regarded as true positive if (1) the predicted human and object bounding box have IoUs larger than 0.5 with corresponding ground-truth boxes,
and (2) the predicted action class is correct.
We computed this metric for each interaction class in HICO-DET and verb class in V-COCO.

Following~\cite{HORCNN_Chao2018WACV, zhang2021upt, zhang2021mining_CDN_NIPS21}, we provide results under \textit{default setting} and \textit{known-object setting} on HICO-DET.
For the first setting, the APs are calculated on the basis of all test images, while the second setting calculate APs based on images that contain the object corresponding to each class.
Under both settings, the result under \textit{full} (a total of 600 interaction classes), \textit{rare} (less than 10 training instances) and \textit{non-rare} (10 or more training instances) classes are all reported.

For V-COCO, we provide results under two evaluation settings: \textit{Scenario 1} and \textit{Scenario 2}.
In the former setting, the detector is required to report an empty box when no object is involved in the interaction, while the object box can be ignored in the latter one.

\begin{table*}[]
    \centering
    \vspace{0.5em}
    \caption{Results on the HICO-DET and V-COCO datasets. The best results are highlighted in \textbf{bold} while the second best ones are \underline{underscored}.}\label{tab:main_results}
    \vspace{-0.5em}
    {
    \begin{tabular}{l C{10ex}C{10ex}C{10ex} C{10ex}C{10ex}C{10ex} cc}
    \toprule
    &     \multicolumn{6}{c}{\textbf{HICO-DET}}     & \multicolumn{2}{c}{\textbf{V-COCO}}  \\
    &     \multicolumn{3}{c}{Default Setting} & \multicolumn{3}{c}{Known-Object Setting} & &  \\
    \cmidrule(lr){2-4} \cmidrule(lr){5-7} \cmidrule(lr){8-9}
    \textbf{Method} &    Full    &   Rare    & Non-rare  &   Full    &  Rare    &   Non-rare    &   Scenario 1  &   Scenario 2  \\
    \midrule
    HO-RCNN~\cite{HORCNN_Chao2018WACV}                   &   7.81    &   5.37    &   8.54    &   10.41 &   8.94    &   10.85   &   -   &   -   \\
    InteractNet~\cite{gkioxari2018detecting_he_CVPR18}   &   9.94    &   7.16    &   10.77   &   -   &   -   &   -   &   40.0    &   -   \\
    GPNN~\cite{qi2018learning_GPNN_ECCV18}               &   13.11   &   9.34    &    14.23   &   -   &   -   &   -   &   44.0    &   -    \\
    TIN~\cite{li2019transferable_TIN_CVPR19}             &   17.03   &   13.42   & 18.11   &   19.17   &   15.51   &   20.26   &   47.8    &   54.2    \\
    DRG~\cite{gao2020drg_ECCV20}                         &   19.26   &   17.74   &   19.71   &   23.40   &   21.75   &   23.89   &   51.0    &   -    \\
    VSGNet~\cite{ulutan2020vsgnet_CVPR20}                &  19.80   &   16.05   & 20.91   &   -   &   -   &   -   &   51.8    &   57.0     \\
    DJ-RN~\cite{li2020djrn_CVPR20}                      &   21.34   &   18.53   & 22.18   &   23.69   &   20.64   &   24.60   &   -   &   -    \\
    PPDM~\cite{liao2020ppdm_CVPR20}                     &   21.94   &   13.97   &   24.32   &   24.81   &   17.09   &   27.12   &   -   &   -   \\
    ConsNet~\cite{liu2020consnet_CVPR20}               &   22.15   &   17.55   &   23.52   &   -   &   -   &   -   &   53.2    &   -    \\
    VCL~\cite{hou2020visual_VCL_ECCV20}               &   23.63   &   17.21   & 25.55   &   25.98   &   19.12   &   28.03   &   48.3    &   -    \\
    ATL~\cite{hou2021affordance_ATL_CVPR21}            &   23.81   &   17.43   & 27.42   &   -   &   -   &   -   &   -   &   -    \\
    IDN~\cite{li2020hoi_IDN}                           &   24.58   &   20.33   & 25.86   &   27.89   &   23.64   &   29.16   &   53.3    &   60.3    \\
    HOTR~\cite{kim2021hotr_CVPR21}                     &   25.10   &   17.34   & 27.42   &   -   &   -   &   -   &   55.2    &   64.4       \\
    FCL~\cite{hou2021detecting_FCL_CVPR21}             &   25.27   &   20.57   & 26.67   &   27.71   &   22.34   &   28.93   &   52.4    &   -     \\
    HOI-Trans~\cite{zou2021end_HOITransformer_CVPR21}   &   26.61   &   19.15   & 28.84   &   29.13   &   20.98   &   31.57   &   52.9    &   -    \\
    AS-Net~\cite{chen2021reformulating_ASNet_CVPR21}     &   28.87   &   24.25   & 30.25   &   31.74   &   27.07   &   33.14   &   53.9    &   -    \\
    SCG~\cite{zhang2021spatially_SCG_ICCV21}            &   29.26   &   24.61   & 30.65   &   32.87   &   27.89   &   34.35   &   54.2    &   60.9    \\
    QPIC~\cite{tamura2021qpic_CVPR21}                &   29.90   &   23.92   & 31.69   &   32.38   &   26.06   &   34.27   &   58.8    &   61.0  \\
    OCN~\cite{yuan2022OCN_HOI}                           &   30.91   &   25.56   &   32.51   &   -   &   -   &   -   &   -   &  - \\
    CDN~\cite{zhang2021mining_CDN_NIPS21}          &   \underline{31.44}   &   \textbf{27.39}   &   \underline{32.64}   &   \underline{34.09}   &   \textbf{29.63}   &   \underline{35.42}   &   \textbf{61.7}    &   \underline{63.8}   \\
    UPT~\cite{zhang2021upt}                            &   \textbf{31.66}   &   \underline{25.94}   & \textbf{33.36}   &   \textbf{35.05}   &   \underline{29.27 } &   \textbf{36.77}   &   \underline{59.0}    &   \textbf{64.5}    \\
    \midrule
    SDT (ResNet-50)    &    \underline{32.45}  & \underline{28.09}  & \underline{33.75}  & \underline{35.95}  & \underline{31.30} & \underline{37.34} &  \underline{60.3} & \underline{65.7}   \\
    SDT (ResNet-101)    &   \textbf{32.97} & \textbf{28.49}   & \textbf{34.31}  & \textbf{36.32} & \textbf{31.90} & \textbf{37.64} &   \textbf{61.8}    & \textbf{67.6}  \\
    \bottomrule
    \end{tabular}
    }
    \vspace{0.5em}
\end{table*}

\vspace{0.5em}
\noindent\textbf{Implementation Details.}
We first fine-tuned DETR for 30 epochs
on the two datasets, which has been pre-trained on MS-COCO\footnote{Note that MS-COCO training set contains some images in the V-COCO test set, which should be excluded in the detector pre-training process ~\cite{tamura2021qpic_CVPR21, zhang2021upt}.}.
Following~\cite{zhang2021upt}, some data augmentation techniques were applied in the detector fine-tuning process:
We scaled the images such that the shorter side is between 480 to 800 pixels while the longer side is at most 1,333 pixels.
Furthermore, each image was cropped with a probability of 0.5 to a random rectangle with each side between 384 to 600 pixels before scaled.
Besides, we also applied color jittering augmentation, where brightness, contrast and saturation are randomly selected between 0.6 to 1.4.
After that, the DETR was frozen in the next stage for interaction recognition.

For each given image, the fine-tuned DETR first perform object detection and generate 100 proposals.
Then, we filter out tokens with confidence score less than 0.2, and keep 3$\sim$15 human/object tokens with higher confidence, based on which our SDT is trained.
We trained the interaction recognition model with the AdamW optimizer~\cite{loshchilov2018decoupled}, which has a learning rate of 2e-4 and weight decay of 1e-4. 
The SDT is trained for 20 epochs and the learning rate is decayed by 10 at the $10$-th epoch.
$L_T$ was set to 3 on both datasets while $L_I$ was 2 and 3 on V-COCO and HICO-DET, respectively.
We set $\lambda$ in Eq.~\ref{eq:interaction_score} to 1 for training and 2.8 for inference.
The token dimension $d$ is set to 256.
For all of the attention mechanism adopted in this paper, we set the number of heads to 8, hidden dimension to 1024 and dropout probability to 0.1.
We conducted all experiments on 4 NVIDIA RTX A5000 GPUs with CUDA 11.1, whereby each GPU has a batch of 4 images, resulting in an effective batch size of 16.
It takes about 6 hours and 40 minutes to train on HICO-DET and V-COCO, respectively.
Besides, we employed two backbone networks, \textit{i.e.}, ResNet-50 and ResNet-101 for global feature extraction, resulting in two variants of SDT.

\subsection{Comparison with State-of-the-art Methods}
\vspace{0.5em}
We compared the proposed SDT with state-of-the-art methods, and reported the results in Table~\ref{tab:main_results}.
It can be observed that, with ResNet-50 as backbone, the proposed method already outperforms existing methods significantly on the two datasets.
For example, on HICO-DET, we surpass UPT~\cite{zhang2021upt} by about $1$ mAP on both \textit{default setting} and \textit{known-object setting}.
On V-COCO, SDT also achieves consistent improvements.
Specifically, under \textit{scenario 1}, our method can outperform the second-best two-stage method UPT~\cite{zhang2021upt} by $1.3$ mAP, while under \textit{scenario 2}, SDT achieves the best performance over all the existing methods. 
Notably, the proposed method can also benefit from a stronger backbone (\textit{i.e.}, ResNet-101). 
For example, on V-COCO, we outperform all existing methods under two scenarios. 
In particular, under \textit{scenario 2}, we can further improve upon SDT with ResNet-50 by $1.9$ mAP, resulting in an improvement of $3.1$ mAP upon the runner-up.
These results prove the superiority of the proposed method over existing approaches.

\begin{figure*}[ht]
    \centering
    \includegraphics[width=1.0\textwidth, clip=0 0 10 0, clip]{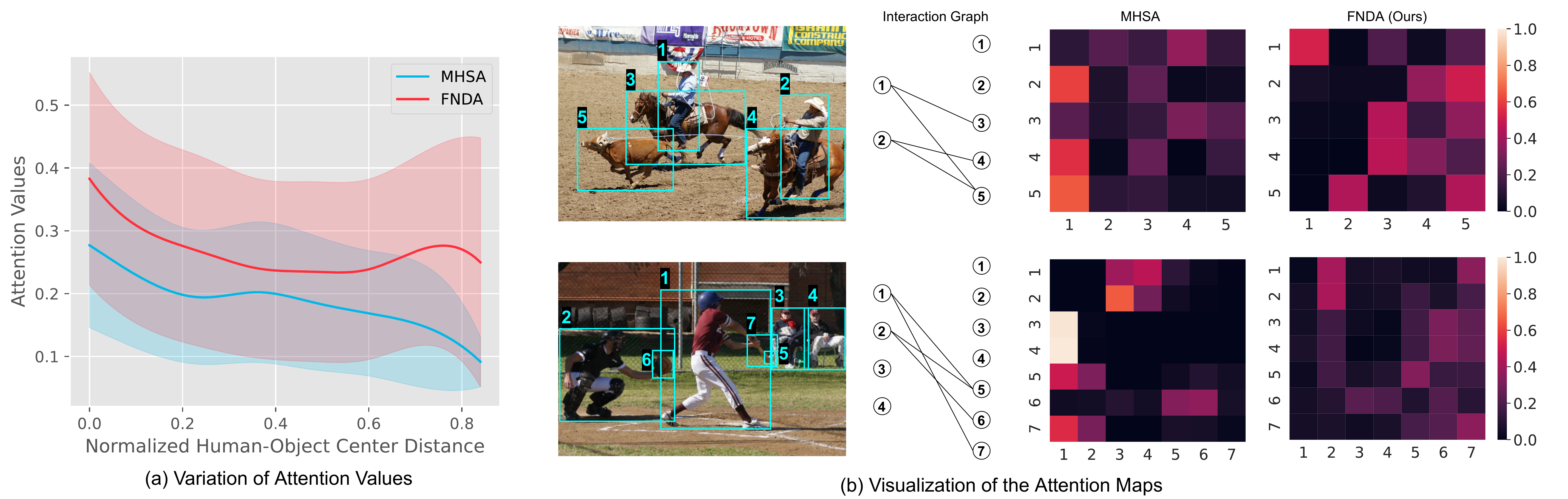}
    \caption{
    Qualitative results from MHSA and the proposed FNDA.
    (a) Attention value variation of MHSA and FNDA with respect to the normalized center distances.
    Note that we split the attention values into bins of 0.05 according to the center distance, where the mean and variance in each bin are both shown. 
    (b) Visualization of attention maps in two randomly selected images. The edges in the interaction graphs indicate the existence of interaction between connected nodes.
    }\label{fig:local_effects}
\end{figure*}

\subsection{Ablation Studies}
\vspace{0.5em}
To further investigate the effects of each component in the proposed method, we conducted extensive ablation studies on the larger HICO-DET dataset with ResNet-50 as backbone.

\begin{table}[]
    \centering
    \caption{Effectiveness of each module in the proposed SDT.}\label{tab:ablation_eachcomponent}
    \vspace{-1em}
    \resizebox{\columnwidth}{!}{%
    \begin{tabular}{ccc|ccc}
         \toprule
         \textbf{T}-Encoder & \textbf{I}-Encoder & DA Loss & Full & Rare & Non-Rare \\
         \midrule
          &  & & 27.08 & 23.01 & 28.30 \\
          \midrule
          \checkmark &  & & 29.09 & 25.06 & 30.29 \\
          & \checkmark &   & 30.04 & 25.32 & 31.45 \\
          &  & \checkmark & 28.09 & 24.04 & 29.29 \\
          \midrule
          \checkmark & \checkmark & & 31.16 & 26.95 & 32.42 \\ 
           \checkmark &  & \checkmark & 30.92 & 27.08 & 32.07 \\
           & \checkmark & \checkmark  & 30.44 & 25.92 & 32.05 \\
           \midrule
           \checkmark & \checkmark & \checkmark & \textbf{32.45} & \textbf{28.09} & \textbf{33.75} \\
         \bottomrule
    \end{tabular}
    }
\end{table}

\vspace{0.5em}
\noindent\textbf{Module Effectiveness.}
We first study the effectiveness of token encoder layers (\textbf{T}-Encoder), interaction encoder layers (\textbf{I}-Encoder) and DA loss, and show the results in Table~\ref{tab:ablation_eachcomponent}.
We can see that all three modules improve the baseline with a clear margin. 
For example, with \textbf{T}-Encoder, our method surpasses the baseline by more than $2$ mAP.
Furthermore, any combination of two modules outperforms the variant with a single module, indicating that the modules are in fact complementary to each other.
In particular, the combination of \textbf{T}-Encoder and \textbf{I}-Encoder can boost upon the baseline by more than $4$ mAP.
Finally, with all of the three modules, the proposed method achieves the best results, which leads to an improvement of more than $5$ mAP upon the baseline.

\vspace{0.5em}
\noindent\textbf{Number of Layers.}
We studied the effects of the number of token encoder layers $L_{T}$ and interaction encoder layers $L_{I}$, and show the results in Table~\ref{tab:num_layers}. 
It can be observed that the combination of $3$ token encoder layers and $3$ interaction encoder layers leads to the best performance.
Increasing $L_T$ or $L_I$ results in less favorable results, while a smaller $L_t$ or $L_I$ also degrades the model performance due to the underfitting problem.

\begin{table}[]
    \centering
    \caption{Performance variation with different numbers of layers.}\label{tab:num_layers}
    \vspace{-1em}
    \begin{tabular}{cc|ccc}
    \toprule
    $L_T$ & $L_I$    & Full  &  Rare & Non-rare  \\
    \midrule
    2     & 2   & 31.06  & 25.98  & 32.58   \\
    2     & 3   & 31.40  & 26.73  & 32.79   \\
    2     & 4   & 31.02  & 25.77  & 32.59   \\
    \midrule
    3     & 2 & 31.71 & 26.95  &  33.13    \\
    3     & 3 & \textbf{32.45} & \textbf{28.09}  &  \textbf{33.75}   \\
    3     & 4 & 31.93 & 27.54  & 33.24     \\
    \midrule
    4     & 2   & 30.91  & 25.01  & 32.67  \\
    4     & 3   & 30.82  & 26.08  & 32.23  \\
    4     & 4   & 30.67  & 25.57  & 32.20  \\
    \bottomrule
    \end{tabular}

\end{table}

\vspace{0.5em}
\noindent\textbf{Quantitative Study on FNDA.}
To demonstrate the superiority of the proposed FNDA, we compared it with the most widely used Multi-Head Self-Attention (MHSA)~\cite{transformer}, and show the results in Table~\ref{tab:fnda}.
We can see that the plain FNDA exceeds MHSA by more than $1$ mAP.
In addition, MHSA also benefits from the token post-processing steps, \textit{i.e.}, ICD and spatial fusion, which also proves the validity of these two.
After combing our FNDA with these two steps, the performance can be further promoted, making it outperform the MHSA by more than 1.5 mAP.

\begin{table}[]
    \centering
    \caption{Comparison between MHSA and FNDA. ++ICD means composing ICD upon spatial fusion.}\label{tab:fnda}
    \vspace{-1em}
    \begin{tabular}{l|ccc}
        \toprule
        Variant &   Full    &   Rare    &   Non-Rare    \\
        \midrule
        MHSA~\cite{transformer}     & 30.66 & 26.06 & 32.03   \\
        +  Spatial Fusion           & 30.80 & 24.78 & 32.60  \\
        ++ ICD       & 30.90 & 26.74 & 32.14   \\
        \midrule
        FNDA (Ours)                 & 31.70 & 26.26 & 33.32   \\
        + Spatial Fusion            & 32.10 & 27.02 & 33.62  \\
        ++ ICD       & \textbf{32.45} & \textbf{28.09} & \textbf{33.75} \\
        \bottomrule 
    \end{tabular}

\end{table}

\begin{figure*}[ht]
    \centering
    \includegraphics[width=1.0\textwidth]{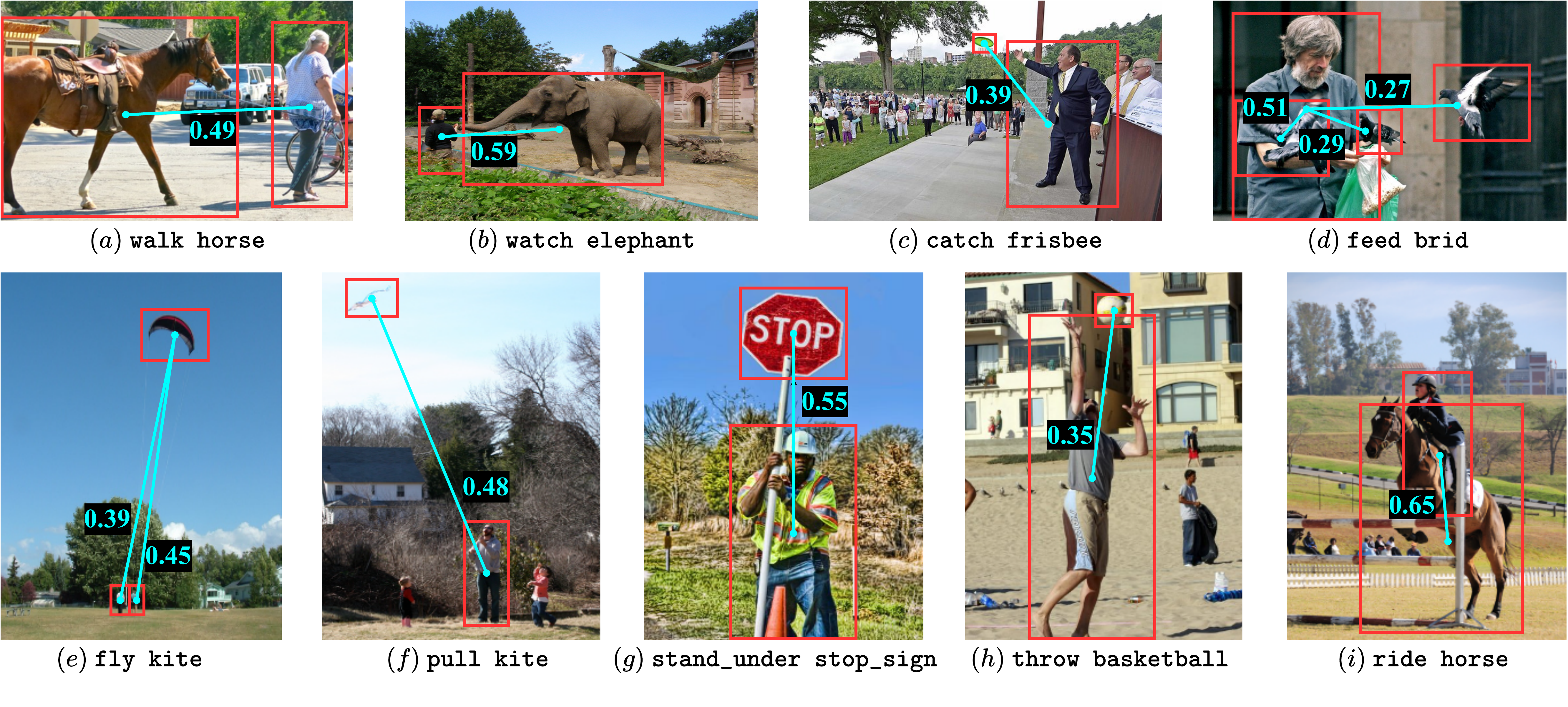}
    \caption{Qualitative results from HICO-DET test set. 
    The number adjacent to the line indicates the predicted score of the target interaction.
    Note that only partial involved human-object interactions are shown for better clarity.
    Best view in color.}
    \label{fig:qualitative_results}
\end{figure*}

\vspace{0.5em}
\noindent\textbf{Qualitative Analysis on FNDA.}
We used all interactive token pairs in the HICO-DET test set, and kept their corresponding attention values in the token encoder layers for both MHSA and our FNDA.
We then illustrate these values with respect to normalized human-object center distances in Fig.~\ref{fig:local_effects} (a).
It can be observed that the attention values generally decline with the increase of distance for both types of attentions, implying that information propagation between distant pairs is more challenging.
More importantly, our FNDA assigns higher attention weights on distant interactions than MHSA, which further justifies the superiority of FNDA over MHSA.

We then used two randomly selected cases to illustrate why FNDA outperforms MHSA.
The two images in Fig.~\ref{fig:local_effects} (b) are selected from HICO-DET test set and the attention map is averaged for far-attention and near-attention.
We observe that the strength of FNDA comes from two aspects:
1) It can effectively propagate information between distant and interactive human-object pairs,
and thus benefit distant interaction recognition.
For example, in the first case of Fig.~\ref{fig:local_effects} (b), the person in box2 is \texttt{lassoing} the \texttt{cow} in box5.
The attention value (2, 5) is minor in MHSA due to the far distance.
By contrast, our FNDA can emphasize more on this human-object pair.
2) FNDA is able to constrain non-interactive human/objects, thereby focusing more on informative ones.
For instance, in the second case of Fig.~\ref{fig:local_effects} (b), the people in box3 and box4 are not involved in any interactions to the people in box1 and box2.
Nonetheless, MHSA assigns higher values in (1, 3), (1, 4), (2, 3), (2, 4), which distracts the people in box1 and box2 from attending to truly related objects.
On the contrary, our FNDA can alleviate this problem by attending to more informative tokens in the image.

\subsection{Qualitative Results}
\vspace{0.5em}
To further understand the effectiveness of the proposed SDT, we randomly selected some images from the HICO-DET test set and visualized SDT prediction results in Fig.~\ref{fig:qualitative_results}.
Note that we only showed partial interactions for clearer view.
It can be observed that our method is capable of handling interaction of diverse distances.
For example, distant instances, such as \texttt{fly} \texttt{kite} in Fig.~\ref{fig:qualitative_results} (e) and \texttt{pull} \texttt{kite} in Fig.~\ref{fig:qualitative_results} (f),
and close instances, such as \texttt{ride} \texttt{horse} in Fig.~\ref{fig:qualitative_results} (i), can both be successfully recognized by the proposed method.
Notably, when the detected object is small (\textit{e.g.}, the \texttt{frisbee} in Fig.~\ref{fig:qualitative_results} (c) and the \texttt{ball} in Fig.~\ref{fig:qualitative_results} (h)), SDT can provide the correct interaction with a high confidence.
Furthermore, it also demonstrates an evident advantage on scenes with complex context, where
multiple objects interact with the same person.
For instance, the \texttt{man} in Fig.~\ref{fig:qualitative_results} (d) is \texttt{feeding} three \texttt{birds} simultaneously, yet our method correctly predicts the interaction for all these three objects.

\section{Conclusion and Future Work}
\vspace{0.5em}
In this work, we propose a novel two-stage method for better modeling distant interactions in HOI detection.
Two fundamental components make our method distinguished from existing approaches. 
First, we design a Far-Near Distance Attention to guide the model attention to flexibly focus on distant and close interactions.
Second, a novel Distance-Aware loss is presented by this paper to handle the long-tailed interaction-distance distribution.
We conduct extensive experiments on two benchmark datasets and observe that our method achieves a new state-of-the-art on them.
Additional analysis further validates that the performance gain is mainly brought by the distant interaction modeling from the proposed method.

As indicated by our empirical findings, a large quantity of distant interactions involve relatively small objects, which impedes the interaction recognition due to the weak detection results.
It is thus promising for HOI detection to be equipped with improved detectors on small-sized objects.
In addition, the 2D image shows limitation in distance estimation as the human-object relations are physically expressed in the 3D world.
In view of this, incorporating the depth information to distance modeling will potentially benefit the learning of distant interactions in HOI detection.


\section*{Acknowledgement}
This research is supported by the National Research Foundation, Singapore under its Strategic Capability Research Centres Funding Initiative. 
Any opinions, findings and conclusions or recommendations expressed in this material are those of the author(s) and do not reflect the views of National Research Foundation, Singapore.

\bibliographystyle{ACM-Reference-Format}
\bibliography{reference}

\end{document}